\documentclass[conference]{llncs}
\usepackage{subfig}
\usepackage{cite}
\usepackage{amsmath,amssymb,amsfonts}
\usepackage{algorithmic}
\usepackage{graphicx}
\usepackage{bm}
\usepackage{textcomp}
\usepackage{float}
\usepackage{xcolor}
\def\BibTeX{{\rm B\kern-.05em{\sc i\kern-.025em b}\kern-.08em
    T\kern-.1667em\lower.7ex\hbox{E}\kern-.125emX}}
\begin{document}

\title{Genetic Bi-objective Optimization Approach to Habitability Score}


\author{Sriram S K \inst{1} \and
Niharika Pentapati \inst{2}}
\institute{Department of Computer Science and Engineering, PES University \email{sriramsk1999@gmail.com} \and Department of Computer Science and Engineering, PES University \email{pniharika369@gmail.com}}

\maketitle

\begin{abstract}
The search for life outside the Solar System is an endeavour of astronomers all around the world. With hundreds of exoplanets being discovered due to advances in astronomy, there is a need to classify the habitability of these exoplanets. This is typically done using various metrics such as the Earth Similarity Index or the Planetary Habitability Index. In this paper, Genetic Algorithms are used to evaluate the best possible habitability scores using the Cobb-Douglas Habitability Score.

Genetic Algorithm is a classic evolutionary algorithm used for solving optimization problems. 
The working of the algorithm is established through comparison with various benchmark functions and its functionality is extended to Multi-Objective optimization. The Cobb-Douglas Habitability Function is formulated as a bi-objective as well as a single objective optimization problem to find the optimal values to maximize the Cobb-Douglas Habitability Score for a set of promising exoplanets. 
\end{abstract}



\section{Introduction}
The search for life has been one of the oldest endeavours of mankind. But only recently have we acquired the capability to take even a step towards this lofty goal. With the first exoplanet discovered in 1991\cite{cochran1991constraints}, we have now reached a point where we have discovered over 4000 exoplanets. We have also taken steps in discovering if life exists on these planets through the use of various metrics such as the Earth Similarity Index\cite{schulze2011two} or the Cobb-Douglas Habitability Score (CDHS) \cite{bora2016cd}. These metrics take various planetary parameters as inputs and give us an intuitive understanding of the likelihood of life existing on these planets. 

The Cobb-Douglas Habitability Production Function (CD-HPF) can quickly give us a score that is representative of the potential of habitability of an exoplanet. It takes in the Radius, Density, Escape Velocity and Mean Surface Temperature of a planet as inputs. All these inputs are in Earth Units(EU) i.e. the metric measurements of these parameters are divided by Earth's own measurements. Simply put, the values of any parameter of Earth in Earth Units is 1.

The Cobb-Douglas function was first developed in 1927\cite{coma1928theory}, seeking to mathematically estimate the relationship between workers, capital and goods produced. In its most standard form for production of a single good with two factors, it is written as $$ Y = AL^\beta K^\alpha $$
Where, $Y$ is the total production, $A$ is total factor productivity, $L$ and $K$ being the labour and capital inputs, and $\alpha,\beta$ being output elasticities of labour and capital respectively. The function itself is highly adaptable and has been utilized for various tasks like revenue models for data centers\cite{saha2016novel}, frameworks for computing scholastic indicators of influence of journals\cite{ginde2016scientobase} successfully.

The CDHS is calculated in a two-fold manner: by calculating the interior-CDHS using radius and density, and the surface-CDHS, by using escape velocity and surface temperature; the final score is computed by a convex combination of the two scores. Thus the function is formulated as a multi-objective optimization problem of the two scores. 

Most optimization functions require the gradient of a function to minimize or maximize it. However, this can prove computationally costly and all functions are not differentiable, and even then, the derivative might not be smooth or continuous. In this paper, we use Genetic Algorithms, a class of gradient-free optimization functions, which are more widely applicable by virtue of them not requiring the derivative of the function to optimize it.

In the book, "On the Origin of Species" by Charles Darwin, he concluded that only those species survived who were successful in adapting to the changing environment and others died. He called this "Natural Selection" which has three main processes; Heredity,Variation and Selection. These involve species receiving properties from their parents, making variations to evolve and then being selected based on their adaptation to the environment for their survival. Along these lines, genetic algorithms \cite{holland1992adaptation} were introduced with five phases of process to solve an optimization problem. We create a initial population of randomly generated elements, known as solutions to the problem and then evaluate the correctness of the solutions using a fitness function which tells us how well the solution helps in optimizing the problem. Genetic Algorithms revolve around the twin principles of Exploration and Exploitation. There must be enough variety in the population to 'explore' the solution space which is usually vast, and on finding good solutions, the algorithm must 'exploit' these solutions and generate incrementally better solutions.

The typical Genetic Algorithm consists of 3 processes: Selection, Crossover and Mutation. In this paper, we use a modified version of a GA that combines the processes of Mutation and Crossover into one. This Proto-Genetic Algorithm is simpler to implement and understand while not compromising on performance.

We evaluate the 'fitness' of the population, that is to say we find the value of the function to be optimized using the members of the population, generate children using \textit{one} parent and then test their fitness as well, choosing the best for the next generation. It is similar to the biological process of asexual reproduction where the child inherits all the traits from one parent alone. In this case, the child is generated from a Gaussian Distribution (as shown in Figure 2) centered at the parent's value.

We illustrate the results of our algorithm on the set of Earth-like exoplanets that is the TRAPPIST system from the exoplanet catalog\cite{phl}, hosted by the Planetary Habitability Laboratory at the University Of Puerto Rico at Acerbio.

\section{Genetic Algorithms}
A Genetic Algorithm (GA) is a meta heuristic which is based on the process of natural selection. It is a subset of the class of Evolutionary Algorithms which take cues from biological processes. They are most commonly used in optimization or search problems as they are capable of searching large combinatorial solution spaces to find globally optimal solutions. 

Figure 1 indicates the pseudo-code of a typical Genetic Algorithm where a \textit{population} of solutions are initialized randomly, given the constraints of a specific problem. The \textit{fitness} of each solution is calculated, which is the value returned by the given function for that solution. Following which the genetic operators of Selection, Crossover and Mutation take place in order to create an incrementally better population. This process is repeated until a termination condition is met, such as a specified number of \textit{generations}. 

\begin{figure}[ht]
    \centering
    \includegraphics[height = 4.5cm, width = 7.5cm]{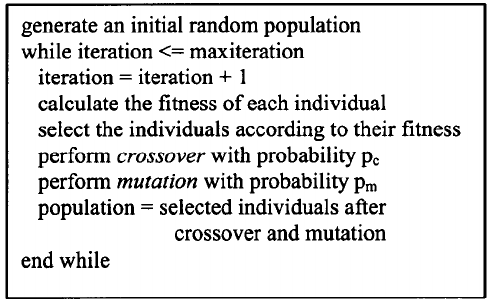}
    \caption{Pseudo-code of a typical Genetic Algorithm}
\end{figure}

\subsection{Proto-Genetic Algorithm}
In this paper, we have utilized a simpler version of the Genetic Algorithm. While GA's typically generate children using traits from both two parents, we have utilized a single-parent reproduction which is both crossover and mutation rolled into one. The best half of the population is selected and a single child is created for each parent. This child is created using a Gaussian Distribution centered at the parent, thus allowing for a mutation of sorts to occur. Due to the nature of the Gaussian Distribution, a small change is much more likely to happen than a drastic one, which reflects real life as well. 

At the end of this process, we have a highly fit population. This algorithm is simpler to understand and implement but gives satisfactory results.   

\begin{figure}
    \centering
    \includegraphics[height = 6.5cm, width = 8.5cm]{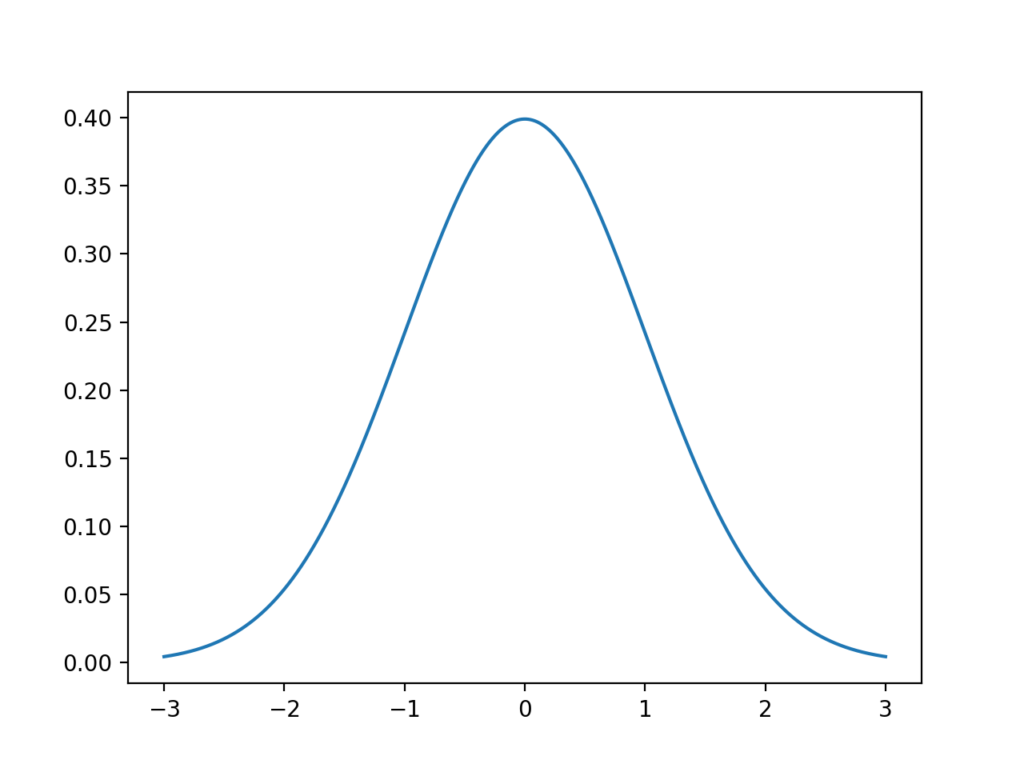}
    \caption{The bell-shaped curve of the Gaussian Distribution}
\end{figure}

\section{Implementation}
\subsection{Single Objective Optimization}
Various test functions like Mishra, Rastrigin, Schaffer and others share many similarities\cite{back1996evolutionary}. All of them have 2 parameters and are highly multimodal. Thus, these functions serve as suitable benchmarks for GA.

We first initialize 2 sets of values for $x$ and $y$, having populations of 200 values each. Our next step is to run the Genetic Algorithm. Here we choose to run the algorithm for 1000 generations, that is to say the processes of crossover, mutation and recombination take place 1000 times at the end of which we have solutions which are very close to the global minimum. The fitness measure here is of course the value of the function for the parameters $x$ and $y$. After calculating the fitness for each pair we then choose the best pairs, i.e. the ones with the lowest fitness and then use them as the parents of the next generation. Generation of children is done using the Gaussian Distribution, allowing us to vary the children slightly in each generation. This is followed by checking the fitness of each child and arranging the children and the parents in order of their fitness. This weeds out all the parents who were not good enough and the children who were worse than their parents. Finally, we remake the population choosing the best of both the old and the new generation.

For example, the Rastrigin Function\cite{muhlenbein1991parallel}, a commonly used benchmark function used to test optimization algorithms due to its highly multimodal nature:
$$ f(x,y) = 20 + x^2 + y^2 - 10(cos(2\pi x) + cos(2 \pi y))$$

\begin{figure}[ht]
    \centering
    \includegraphics[height = 5cm, width = 8cm]{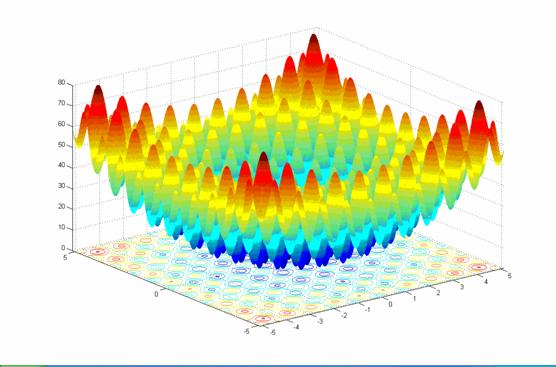}\\
    \caption{The Rastrigin Function}
\end{figure}

The Rastrigin function in Figure 3 has a global minima of 0.0 with the domain being from -5.12 to 5.12. Thus, our algorithm generates 200 values of $x$ and $y$ in the given domain, which is the first generation of the algorithm. They are sorted according to their fitness and new children are generated from the best half of the population. Following this the population is remade by sorting according to fitness again and the second generation is created, with the members being slightly \textit{fitter} than their parents. Table 1 compares the actual global minima and that obtained using GA for various test functions.

\begin{table}[htbp]
\caption{Single Objective Optimization Results}
\begin{center}
\begin{tabular}{|l|c|c|}
\hline
\textbf{Test}&\multicolumn{2}{|c|}{\textbf{Global Minimum}} \\
\cline{2-3} 
\textbf{Functions} & \textbf{\textit{Actual Values}}& \textbf{\textit{GA Values}} \\
\hline
Easom & -1 & -0.999 \\
\hline
Rastrigin & 0.0 & 0.0003 \\
\hline
Ackley & 0.0 & 0.009 \\
\hline
Beale & 0.0 & 0.0 \\
\hline
Goldstein-Price & 3.0 & 3.0001 \\
\hline
Mishra No.4 & -0.199 & -0.193 \\
\hline
Cross-in-tray & -2.06 & -2.06 \\
\hline
Eggholder & -959.64 & -959.27 \\
\hline
Holder table & -19.208 & -19.208 \\
\hline
McCormick & -1.913 & -1.913 \\
\hline
Schaffer No.4 & 0.292 & 0.292 \\
\hline
\multicolumn{3}{l}{$^{\mathrm{}}$}
\end{tabular}
\label{tab1}
\end{center}
\end{table}

\subsection{Constrained Optimization}
These functions are also single objective optimization problems, however they are constrained. Whereas the previous batch of functions are only limited by the search domains, these functions have additional constraints. They tend to be more challenging to optimize. 

Mishra's Bird Function displayed in Figure 4 \cite{mishra2006some} has a global minima of -106.76 and the domain being from -10 to 0 for $x$ and -6.5 to 0 for $y$. The function is given as:
\begin{multline*}
    f(x,y) = sin(y).e^{((1-cos(x))^2)} + cos(x).e^{((1-sin(y))^2)} + (x-y)^2
\end{multline*}

In addition to minimizing this, the solutions must also not violate the additional constraint which is:

$$ (x+5)^2 + (y+5)^2 < 25 $$

We follow the same procedure as with single objective optimization, albeit making sure the solutions do not violate the constraints along with the upper and lower bounds of the domain. Children generated will be discarded if they do not satisfy the constraints. The standard deviation of the Gaussian Distribution goes on increasing to widen the search range if a large number of solutions are discarded. Table 2 lists different test functions along with their actual and GA obtained global minimum.

\begin{figure}[ht]
    \centering
    \includegraphics[height = 7cm, width = 8cm]{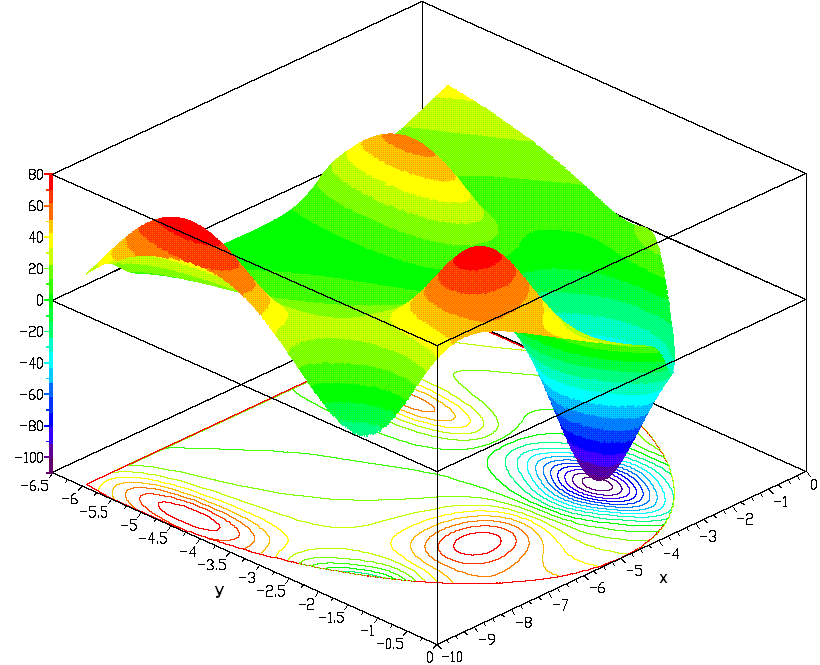}
    \caption{Mishra's Bird Function (Constrained)}
\end{figure}

\begin{table}[htbp]
\caption{Constrained Optimization Results}
\begin{center}
\begin{tabular}{|l|c|c|}
\hline
\textbf{Test}&\multicolumn{2}{|c|}{\textbf{Global Minimum}} \\
\cline{2-3} 
\textbf{Functions} & \textbf{\textit{Actual Values}}& \textbf{\textit{GA Values}} \\
\hline
Rosenbrock (with a cubic and a line) & 0.0 & 0.0009 \\
\hline
Rosenbrock (disk) & 0.0 & 0.0 \\
\hline
Mishra's Bird & -106.76 & -106.76 \\
\hline
Townsend  & -2.02 & -2.02 \\
\hline
Simionescu & -0.072 & -0.0719 \\
\hline
\multicolumn{3}{l}{$^{\mathrm{}}$}
\end{tabular}
\label{tab2}
\end{center}
\end{table}

\begin{figure*}
    \centering
    \includegraphics[height = 6cm, width = 13cm]{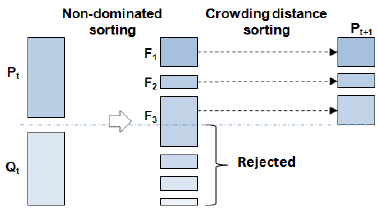}
    \caption{Basics of NSGA-II Procedure}
\end{figure*}

\subsection{Multi-Objective Optimization}
Whereas for single-objective optimization problems there exists a single solution which is the best value, no such solution exists for non-trivial multi-objective problems. 

Multi-objective Optimization problems involve minimizing/maximi\-zing more than one function simultaneously. If these functions are competitive i.e. minimizing one maximizes the other, it is not possible to find a single best solution. Instead we get a set of \textbf{non-dominating} solutions known as a \textbf{Pareto Front} (as shown in Figure 6). In the absence of other information, each solution in the Pareto Front is equally valid and no solution can be said to be better than another. 

Pareto fronts are based on the idea of dominance. If $\vec{x},\vec{y}$ are two solutions, then $\vec{x}$ is said to dominate $\vec{y}$ if 
$$f_i(\vec{x})) \leq f_i(\vec{y})) \ \forall \ i = 1, 2, 3, ...k $$

In another words, the vector $\vec{x}$ is said to dominate $\vec{y}$ if and only if, $f(x) \leq f(y)$ for every single objective in the multi-objective optimization problem. We say that a vector of decision variables $\vec{x} \in F$ is said to be Pareto optimal if no other vector $\vec{x} \in F$ exists such that $f(\vec{y}) \leq f(\vec{x}).$ A multi-objective optimization consists of finding the best Pareto front for a given set of objectives. 

There are various algorithms for multi-objective optimization. Indeed, one such algorithm, Particle Swarm Optimization (PSO) has already been used in solving the CD-HPF\cite{theophilus2018novel}. PSO have many advantages over GA\cite{eberhart1998comparison} and hybrid PSO-GA have also been used in problems like intelligent routing\cite{v2015qos} to great success. 

When solving a multi-objective optimization problem using GA, a different approach must be taken. While in single objective problems we can directly compare function values as fitness and choose the best parents, the same cannot be done when we have multiple objectives to optimize. There are numerous algorithms such as MOGA\cite{fonseca1993genetic}, NSGA\cite{srinivas1994muiltiobjective}.In this paper, we have used one of the most popular multi-objective optimization algorithms, NSGA-II\cite{deb2002fast}.

\begin{figure}[H]
    \centering
    \includegraphics[height = 7cm, width = 8cm]{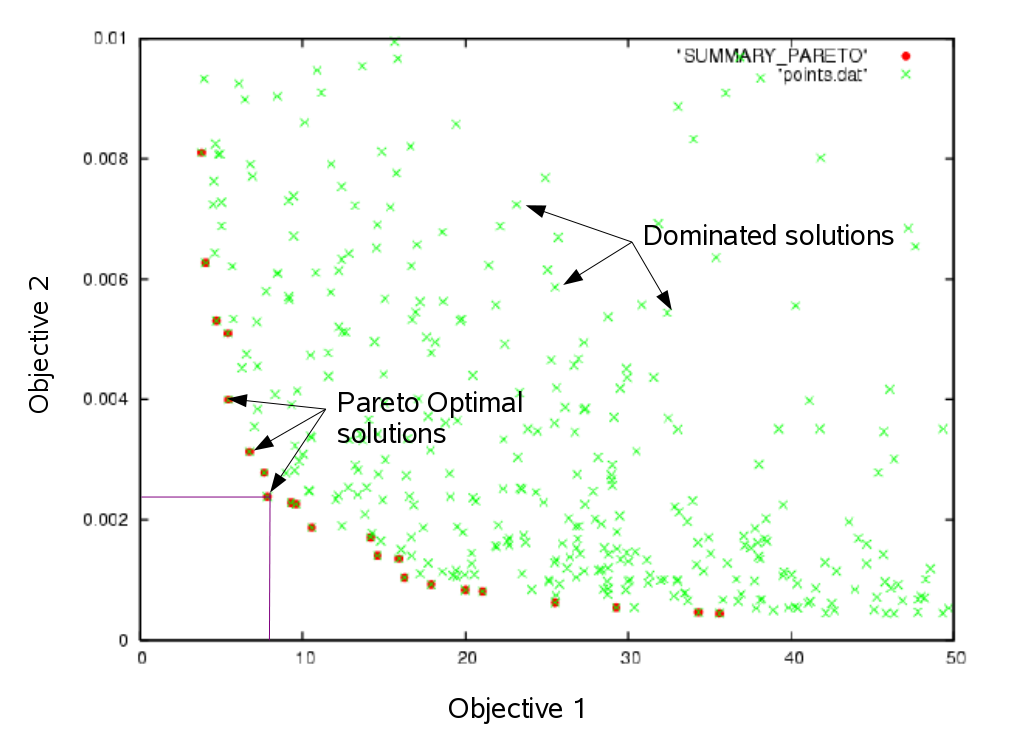}
    \caption{A Pareto Front}
\end{figure}

NSGA-II as illustrated in Figure 5 consists of two new processes in order to assign fitness to solutions. The first is the \textbf{non-dominated sort}, where the solutions are sorted into sets of non-dominating solutions i.e. fronts. The second is the \textbf{crowded-comparison}, which ensures that solutions which have fewer number of solutions in their vicinity have a higher chance of getting selected. In other words, this algorithm favours non-dominated solutions which are well distributed.

Thus we can assign a fitness to the solutions even with multiple objectives. Following this we use our proto-genetic algorithm to evolve the chosen solutions and continue the process iteratively until we have our population closely resembling the optimal Pareto Front. Figures 7 and 8 compare the actual and obtained pareto fronts of different test functions.

\begin{figure}
    \centering
    \subfloat[Poloni]{\includegraphics[width=4cm]{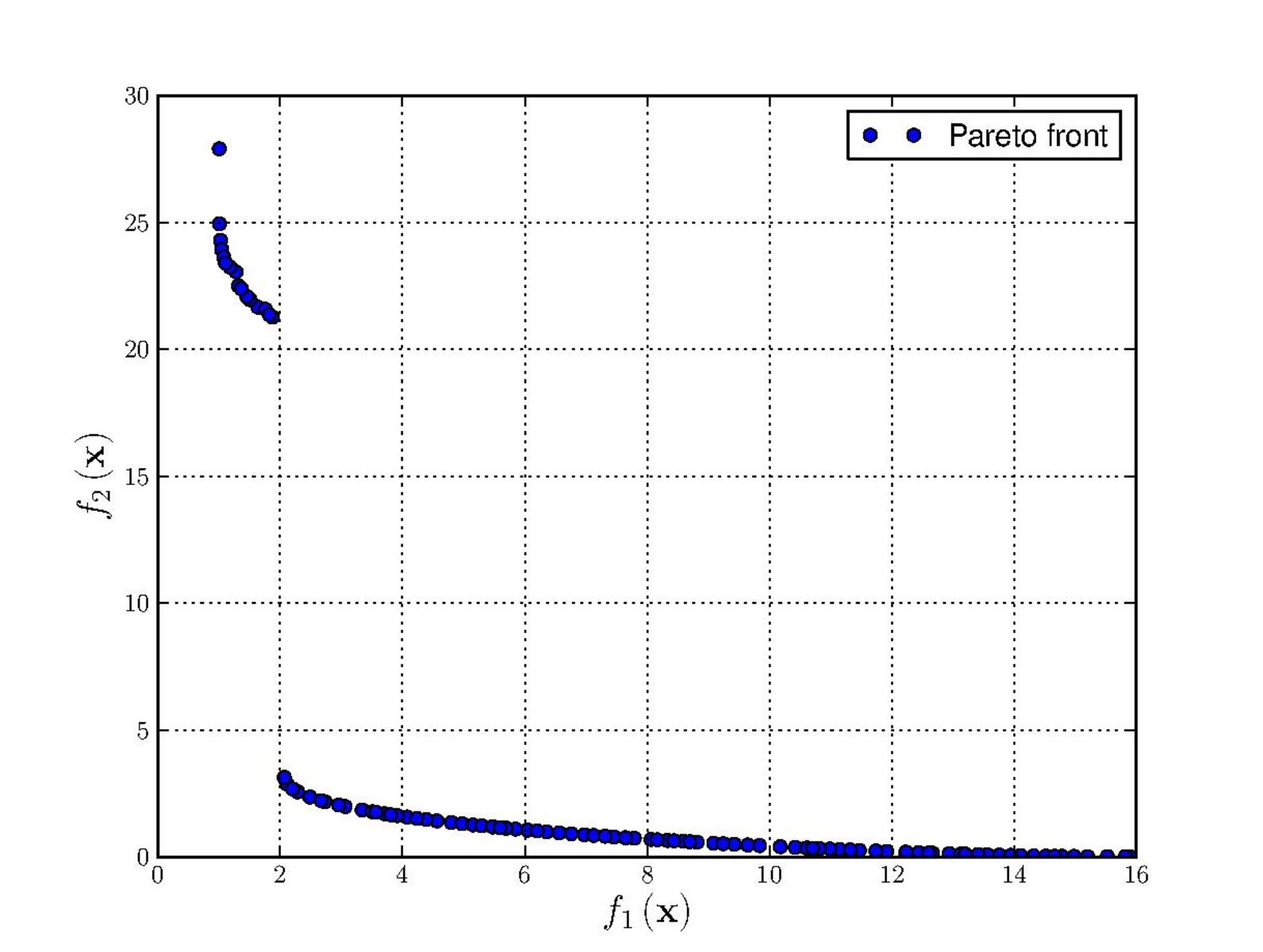}}
    \qquad
    \subfloat[Schaffer1]{\includegraphics[width=4cm]{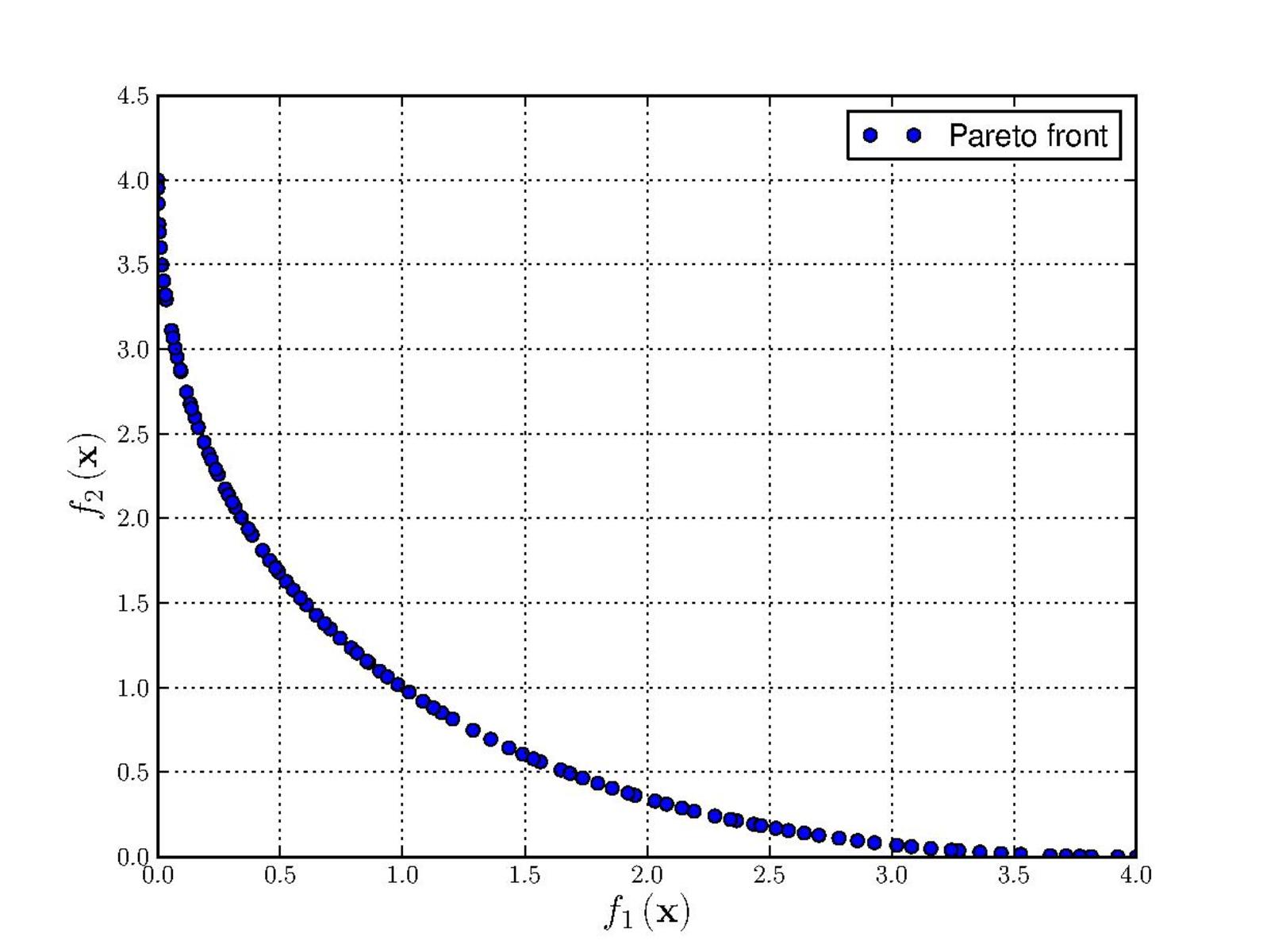}}
    \qquad
    \subfloat[CTP 1]{{\includegraphics[width=4cm]{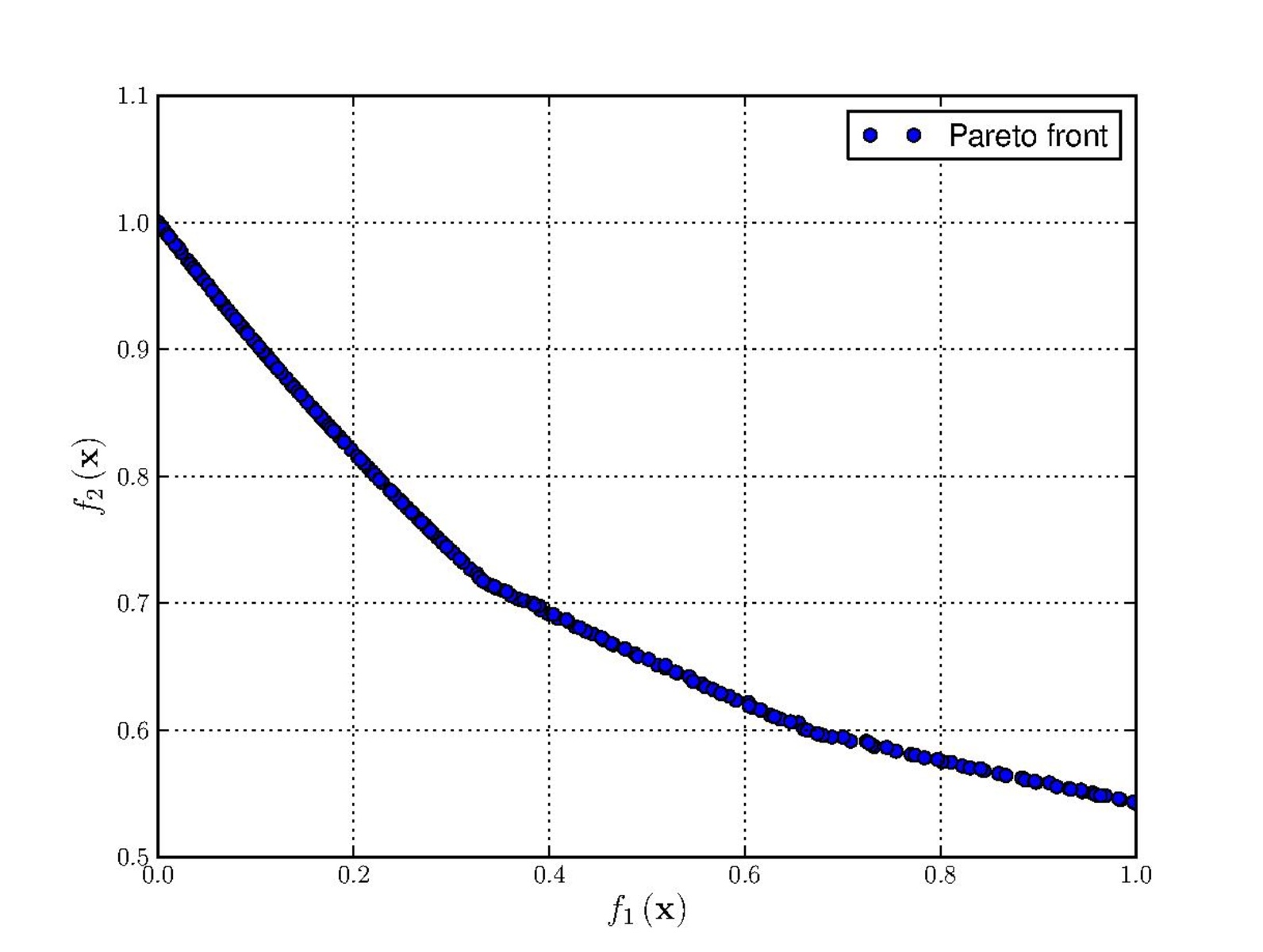}}}
    \qquad
    \subfloat[Constr-Ex]{{\includegraphics[width=4cm]{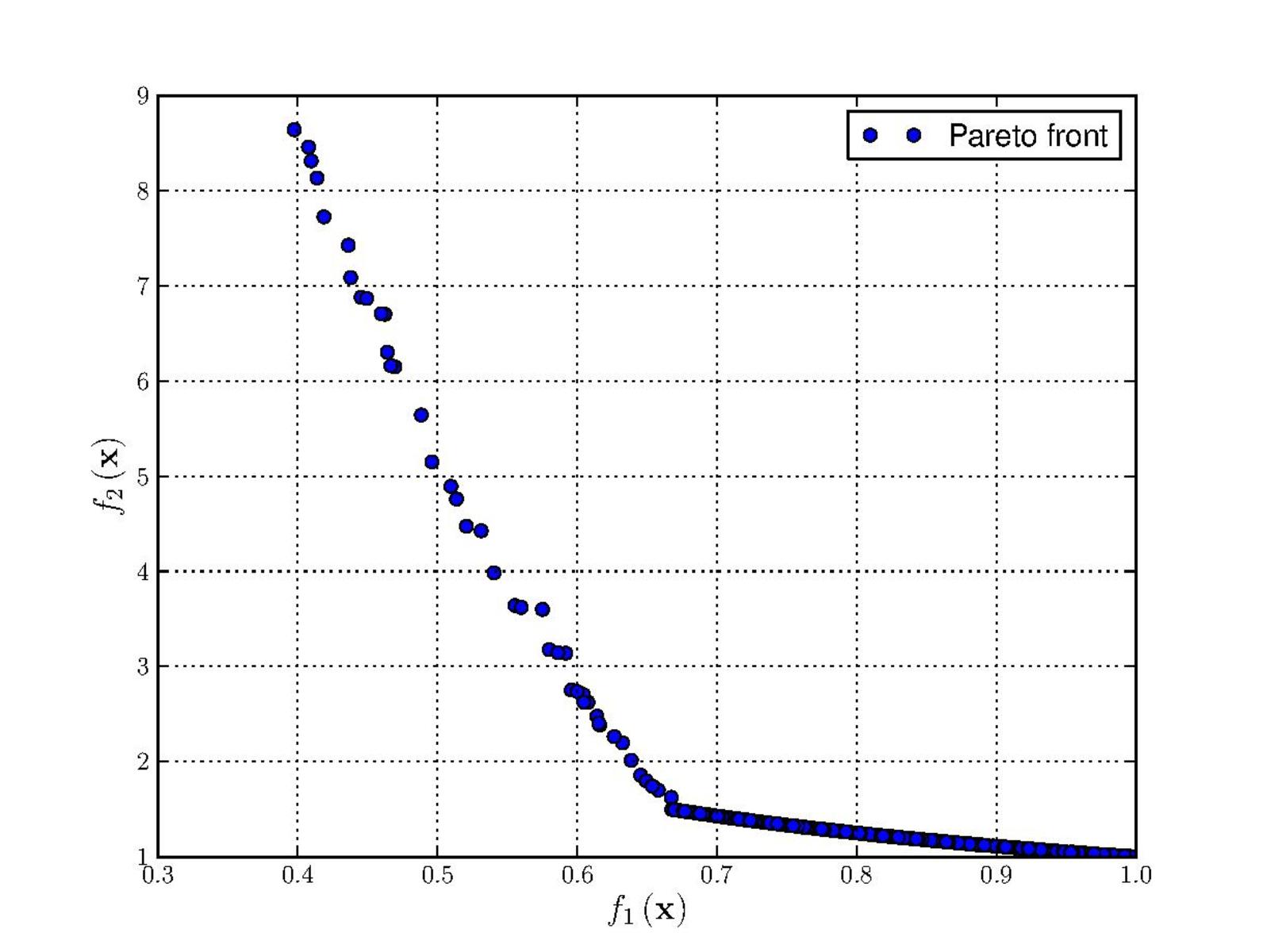}}}
    \qquad
    \subfloat[Binh and Korn]{{\includegraphics[width=4cm]{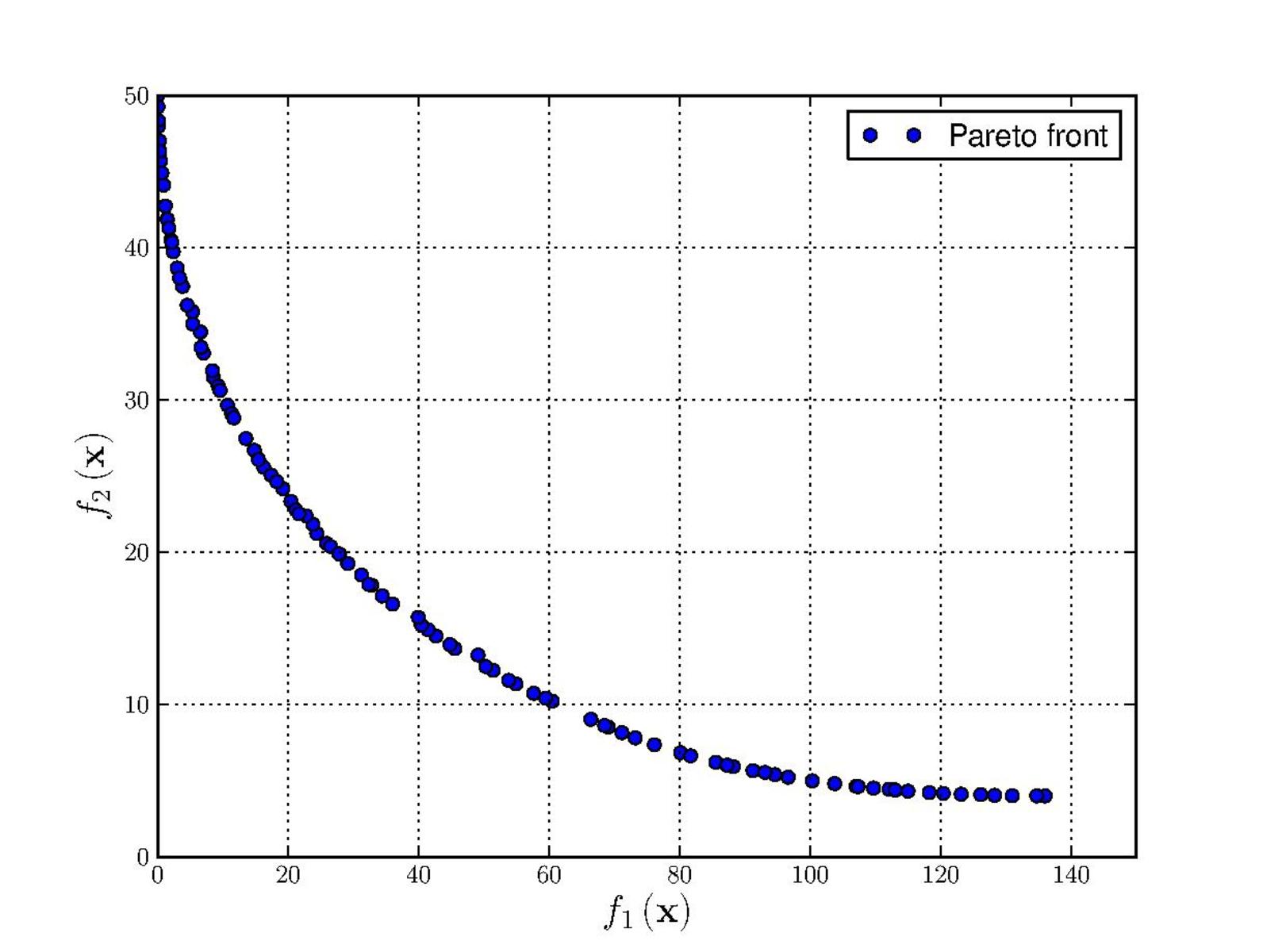}}}
    \qquad
    \subfloat[Chakong and Haimes]{{\includegraphics[width=4cm]{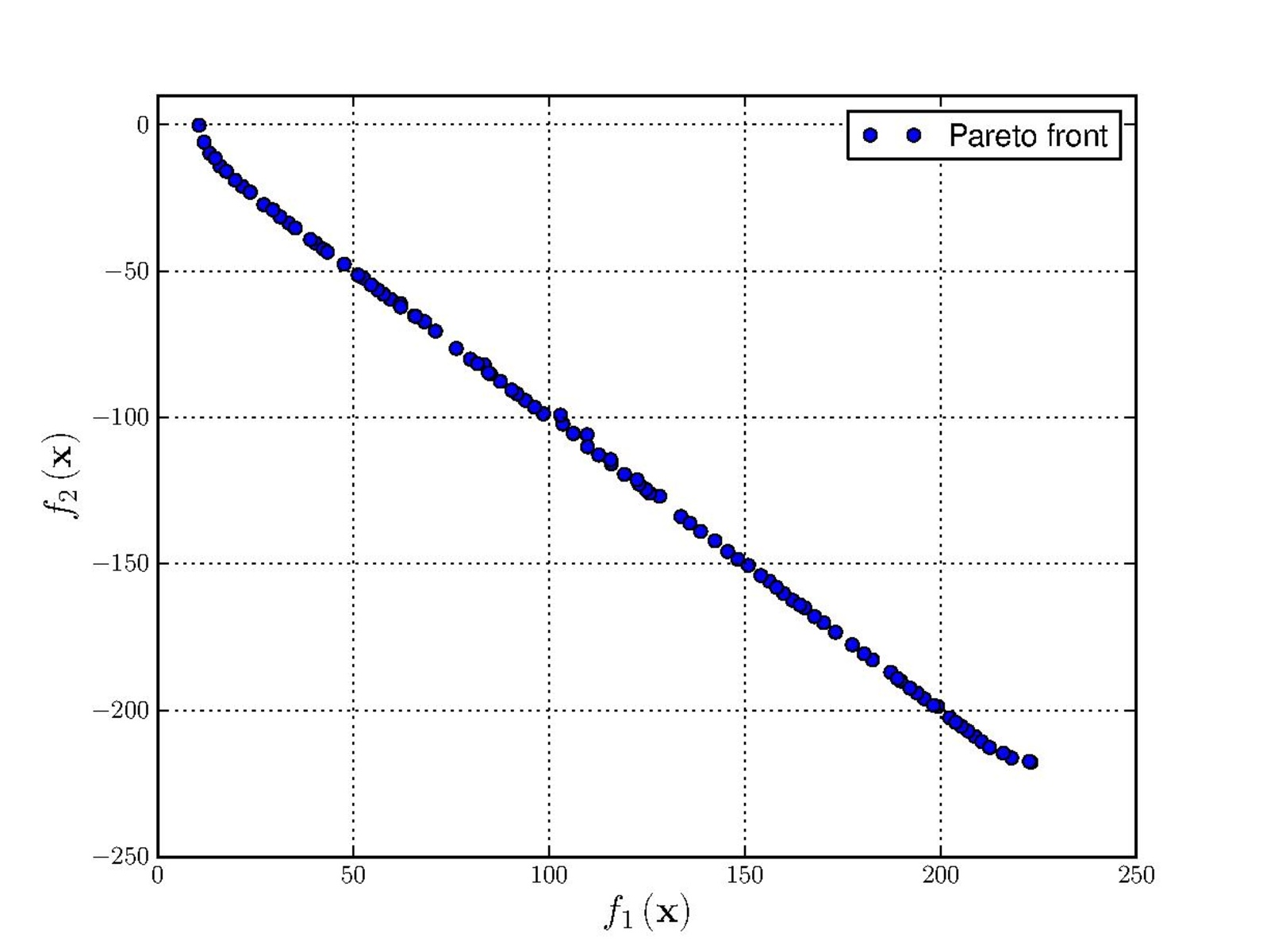}}}
    \caption{Actual Pareto Fronts of Test Functions}
    \label{fig:paretowiki}
\end{figure}
\begin{figure}[ht]
    \centering
    \subfloat[Poloni]{\includegraphics[width=4cm]{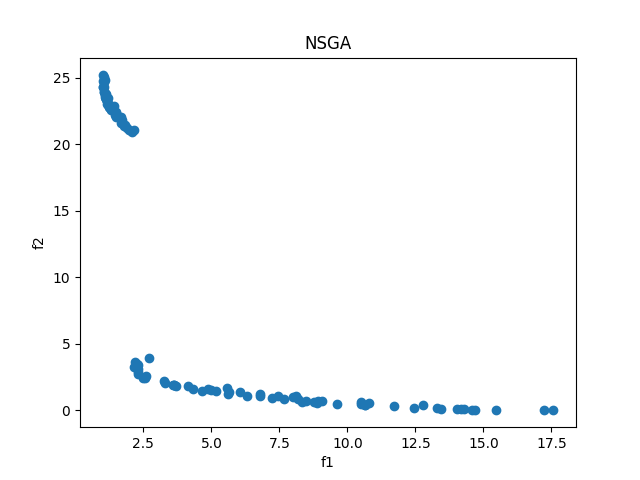}}
    \qquad
    \subfloat[Schaffer1]{\includegraphics[width=4cm]{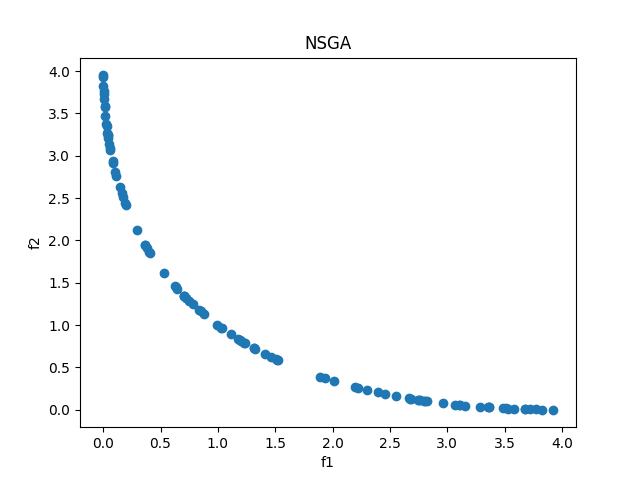}}
    \qquad
    \subfloat[CTP 1]{{\includegraphics[width=4cm]{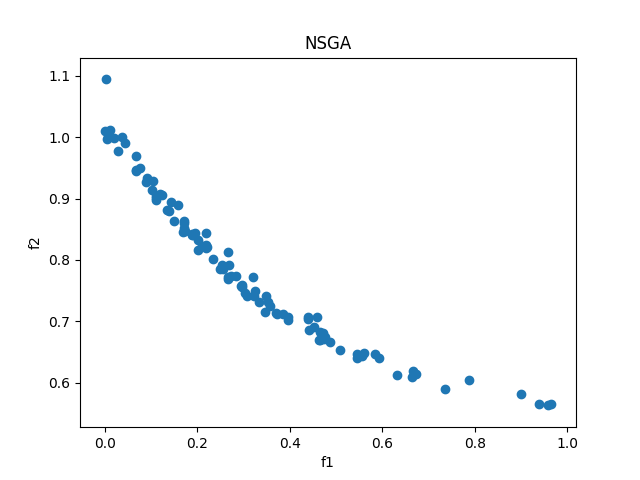}}}
    \qquad
    \subfloat[Constr-Ex]{{\includegraphics[width=4cm]{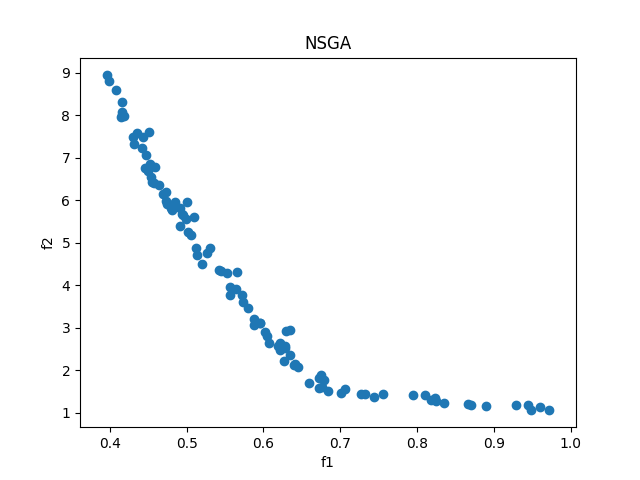}}}
    \qquad
    \subfloat[Binh and Korn]{{\includegraphics[width=4cm]{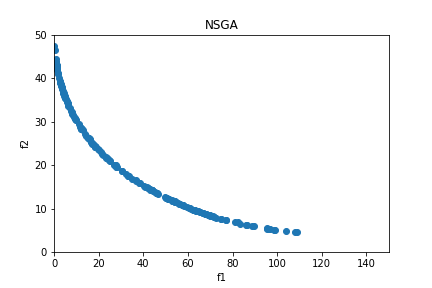}}}
    \qquad
    \subfloat[Chakong and Haimes]{{\includegraphics[width=4cm]{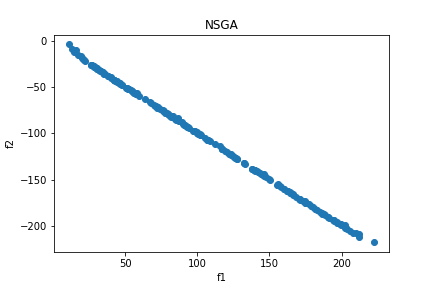}}}
    \caption{Obtained Pareto Fronts}
    \label{fig:pareto}
\end{figure}

\section{Cobb-Douglas Habitability Function}
The Cobb-Douglas Habitability Function is given as follows:
\begin{equation*}
     Y = R^\alpha.D^\beta.V_e^\delta.T_s^\gamma
\end{equation*}

It can also be formulated as a bi-objective optimization problem for easy visualization and understanding. The Cobb-Douglas Habitability Score ($Y$) is divided into two components, CDHS-interior ($Y_i$) and CDHS-surface ($Y_s$). The CDHS is estimated by maximizing both $Y_i$ and $Y_s$ which are defined as follows:
\begin{equation*}
     Y_i = R^\alpha.D^\beta
\end{equation*}
\begin{equation*}
     Y_s = V_e^\delta.T_s^\gamma
\end{equation*}
These functions are subject to the constraints:
\begin{equation*}
    \alpha + \beta \leq 1
\end{equation*}
\begin{equation*}
     \delta + \gamma \leq 1
\end{equation*}
\begin{equation*}
     0 < \alpha , \beta , \delta , \gamma < 1
\end{equation*}

Where $\alpha,\beta,\gamma,\delta$ are the \textit{elasticities} of the planetary parameters Radius, Density, Escape Velocity and Mean Surface Temperature. The quality of this model is well noted.\cite{saha2018theoretical}

Thus we have a bi-objective optimization problem where we have to optimize $CDHS_i$ and $CDHS_s$ simultaneously.

\begin{equation*}
     max f(\vec{x}) = [Y_i,Y_s]
\end{equation*}

However, since $V_e = \sqrt{\dfrac{2GM}{R}}$, we know that increasing surface score is not possible without compromising on interior score and vice versa. Thus, as shown in \cite{bora2016cd}, we use the following relationships:

$$V_e = \dfrac{\delta}{\alpha}\dfrac{W_R}{W_{V_e}}R $$

Where $W_R$ and $W_{V_e}$ are weights of $R$ and $V_e$ respectively. Rearranging the equation we get:

$$ \delta = \alpha \dfrac{V_e}{R} C$$ 

where, 

$$ C = \dfrac{W_{V_e}}{W_R} $$

In order to bring out the trade-off between the two components of the Cobb-Douglas Habitability Score, we calculate $\delta$ from the other parameters, optimizing the variables $\alpha,\beta,\gamma$ and $C$.

We apply the aforementioned proto-genetic algorithm modified with NSGA-II on a set of planets from the exoplanet catalog hosted by the Planetary Habitability Laboratory at the University Of Puerto Rico, the TRAPPIST system. The results are shown in Table 3 with illustrations in Figure 9. 

\section{Results}
After testing on multiple exoplanets in the catalog, we found promising results similar to that of past approaches\cite{saha2018theoretical}. 

\begin{table}[htbp]
\caption{Comparison of CDHS using GA with past approaches}
\begin{center}
\begin{tabular}{|l|c|c|}
\hline
\cline{2-3} 
\textbf{Exoplanets} & \textbf{CDHS(2018)}& \textbf{CDHS(GA)} \\
\hline
TRAPPIST-1 b & 1.0410 & 1.3753 \\
\hline
TRAPPIST-1 c & 1.1589 & 1.2073 \\
\hline
TRAPPIST-1 d & 0.8870 & 1.0146 \\
\hline
TRAPPIST-1 e & 0.9093 & 0.9990 \\
\hline
TRAPPIST-1 f & 0.9826 & 1.0389 \\
\hline
TRAPPIST-1 h & 0.8025 & 0.9973 \\
\hline
Proxima Cen b & 1.08297 & 1.11909 \\
\hline
\multicolumn{3}{l}{$^{\mathrm{}}$}
\end{tabular}
\label{tab3}
\end{center}
\end{table}

\begin{figure}[ht]
    \centering
    \subfloat[TRAPPIST-b]{\includegraphics[width=4cm]{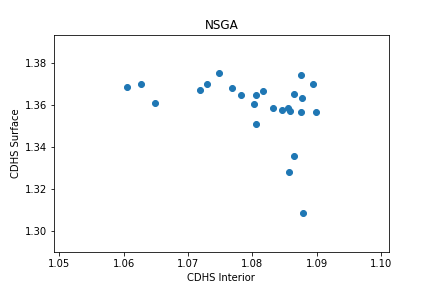}}
    \qquad
    \subfloat[TRAPPIST-c]{\includegraphics[width=4cm]{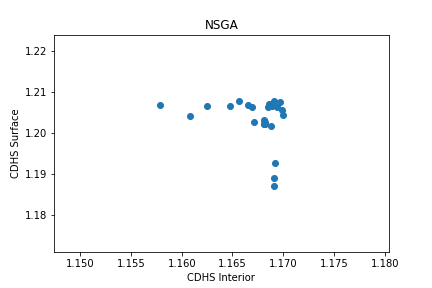}}
    \qquad
    \subfloat[TRAPPIST-d]{{\includegraphics[width=4cm]{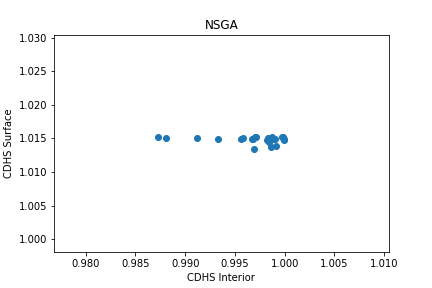}}}
    \qquad
    \subfloat[TRAPPIST-e]{{\includegraphics[width=4cm]{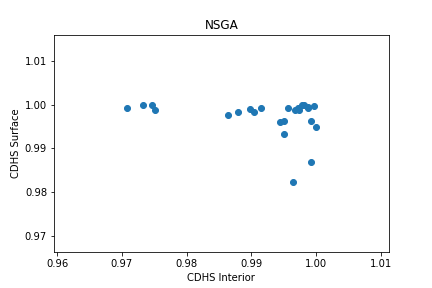}}}
    \qquad
    \subfloat[TRAPPIST-f]{{\includegraphics[width=4cm]{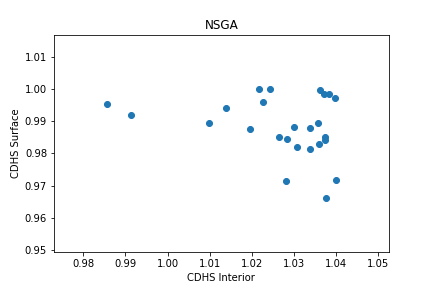}}}
    \qquad
    \subfloat[TRAPPIST-h]{{\includegraphics[width=4cm]{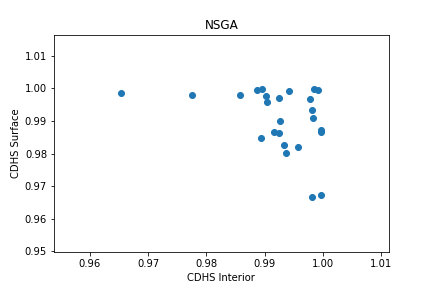}}}
    \caption{The TRAPPIST system of exoplanets}
    \label{fig:trappist}
\end{figure}

The Pareto fronts also show a trend where increase in one score is compensated for by decrease in the other. These complementary scores bring out the trade-off between $Y_i$ and $Y_s$.  

The final score is calculated as the weighted linear combination of interior and surface score where the weights sum up to 1.

\begin{equation*}
     Y = w_i.Y_i + w_s.Y_s
\end{equation*}
\begin{equation*}
     w_i + w_s = 1
\end{equation*}

We set the weights $w_i$ and $w_s$ as 0.5 i.e. equal weights. Thus the calculated CDHS is the mean of the surface score and the interior score. With different weight pairs we get a range of habitability scores for each planet instead of a hard score, making the model more robust than other metrics.

In order to ensure the results are consistent, the CD-HPF was also solved as a single objective optimization problem i.e. its original form:

\begin{equation*}
     Y = R^\alpha.D^\beta.V_e^\delta.T_s^\gamma
\end{equation*}

Similar to other single-objective optimization problems, we generated populations of $\alpha, \beta, \gamma$ and $\delta$ and evolved them with the proto-genetic algorithm. The results were similar and establish the veracity of the multi-objective optimization approach as listed in Table 4.

\begin{table}[htbp]
\caption{CDHS obtained using Multi-Objective and Single Objective Optimization}
\begin{center}
\begin{tabular}{|l|c|c|}
\hline
\cline{2-3} 
\textbf{Exoplanets} & \textbf{CDHS(Multi-Objective)}& \textbf{CDHS(Single Objective)} \\
\hline
TRAPPIST-1 b & 1.3753 &  1.3684 \\
\hline
TRAPPIST-1 c & 1.2073 &  1.2065\\
\hline
TRAPPIST-1 d & 1.0146 &  1.0138\\
\hline
TRAPPIST-1 e & 0.9990 &  0.9972\\
\hline
TRAPPIST-1 f & 1.0389 &  1.0343\\
\hline
TRAPPIST-1 h & 0.9973 &  0.9929\\
\hline
Proxima Cen b & 1.11909 &  1.1158\\
\hline
\multicolumn{3}{l}{$^{\mathrm{}}$}
\end{tabular}
\label{tab4}
\end{center}
\end{table}

\section{Conclusion}

In this paper, we have used a Proto-Genetic algorithm along with NSGA-II to calculate the best habitability scores for different exoplanets using the Cobb-Douglas Habitability Production Function. The optimizing capability of the proto-genetic algorithm was well established by testing on numerous benchmark functions of the single objective, constrained single-objective and multi-objective optimization types. Finally the algorithm was applied in calculating the habitability scores of promising exoplanets from the TRAPPIST system. The results were further verified using the single-objective optimization approach as well, establishing the merit of a genetic bi-objective optimization approach to habitability scores.

\section*{Acknowledgment}

We thank our teachers who gave us this opportunity to work on something innovative and provided us a rich learning experience in the process. We thank Dr. Snehanshu Saha for assistance with this endeavour and guiding us in this project. And finally, we would also like to show our gratitude to the PES University for allowing us to go beyond our coursework and work on a project with practical applications. 

\bibliographystyle{splncs04}
\bibliography{bibliography.bib}

\end{document}